\pdfoutput=1

\documentclass[11pt]{article}

\usepackage{acl}

\usepackage{times}
\usepackage{latexsym}
\usepackage{amsmath}
\usepackage{array}
\usepackage{comment}
\usepackage{multirow}

\usepackage[T1]{fontenc}

\usepackage[utf8]{inputenc}

\usepackage{microtype}

\usepackage{graphicx}
\graphicspath{ {images/} }

\newcommand{\rmb}[1]{}
\newcommand{\jmf}[1]{} 

\title{Automatic Identification of Motivation for Code-Switching in Speech Transcripts}

\author{Ritu Belani \\
  The Harker School \\
  \texttt{ritubelani@gmail.com} \\
  \And
  Jeffrey Flanigan \\
  University of California, Santa Cruz \\
  \texttt{jmflanig@ucsc.edu} \\}

\begin{document}
\maketitle
\begin{abstract}

Code-switching, or switching between languages, occurs for many reasons and has important linguistic, sociological, and cultural implications. Multilingual speakers code-switch for a variety of purposes, such as expressing emotions, borrowing terms, making jokes, introducing a new topic, etc.  The reason for code-switching may be quite useful for analysis, but is not readily apparent.  To remedy this situation, we annotate a new dataset of motivations for code-switching in Spanish-English. We build the first system (to our knowledge) to automatically identify a wide range of motivations that speakers code-switch in everyday speech, achieving an accuracy of 75\% across all motivations. Additionally, we show that the system can be adapted to new language pairs, achieving 66\% accuracy on a new language pair (Hindi-English), demonstrating the cross-lingual applicability of our annotation scheme.

\end{abstract}

\section{Introduction}

\rmb{ready to edit}
Code-switching, or switching between languages within the same utterance or sentence~\cite{poplack1980sometimes},
commonly emerges in conversations between multilinguals and in written communication such as social media. In today’s intersecting multilingual world, it is essential to develop computational tools that can process and analyze code-switched speech and text.

In recent years, there has been much progress in processing code-switched language. Many code-switched datasets have been collected\jmf{citations}.
Code-switching datasets which have been made widely available extend to areas of natural language inference~\cite{khanuja2020new}, semantic parsing~\cite{einolghozati2021volumen}, language identification~\cite{king2013labeling, molina2019overview}, POS tagging~\cite{solorio2008part, jamatia2015part, singh2018twitter}, NER~\cite{gupta2016hybrid, aguilar2019named}, sentiment analysis~\cite{vilares2015sentiment, joshi2016towards}, conversational systems~\cite{banerjee2018dataset, chandu2019code}, and machine translation~\cite{dhar2018enabling, menacer2019machine}. Workshops held on computational approaches to code-switching created shared tasks on language identification~\cite{solorio2014overview} and Named Entity Recognition (NER)~\cite{aguilar2019named} in code-switched texts. Nuanced tasks like humor detection, sarcasm detection, and hate detection have been applied to Hindi-English code-switched data~\cite{bansal2020code}.

Despite these achievements, there is relatively little work on identifying the motivations for code-switching.  Although there are annotations schemes~\cite{zentella1998growing, hartmann2018integrated} \jmf{are there others?} and some annotated datasets~\cite{dey2014hindi, begum2016functions, lee2015emotion, rudra2019identifying}\jmf{are there others?}, to our knowledge, there is no work automatically identifying the communicative motivation behind a code-switch across the full range of motivations~\cite{zentella1998growing}.\jmf{we have a qualifier ``in the way that linguistics research..." Are there works that identify communicative motivation behind code-switching that don't satisfy that qualifier? If so, we should talk about them in the related work}\rmb{yes, see Rudra's work identifying whether the code-switch is opinionated or not, and Lee and Wang's work identifying just the emotions in the code-switch}\jmf{can you add those citations to the related work?}

We believe there are many applications for the task we propose in this paper, including identification of speaker sentiment, detecting topic change in discourse analysis, lexical analysis of borrow words, and cross-lingual entity recognition. As an example application, our motivation is to reduce the misdiagnosis of bilingual students with language disorders by identifying the reason behind their code-switching. Despite code-switching being a natural practice for bilinguals, one misconception is that it indicates confusion between two languages and a language deficit~\cite{national2017promoting}. The system we propose would effectively assist speech therapists to make a more accurate diagnosis of Hispanic bilingual children by allowing monolingual speech therapists to receive the same depth of information from code-switching that bilingual speech therapists, who are familiar with the syntax of both languages, would receive.
\jmf{We can use this as a case study in applications, but it would be good to add more possible applications.  Perhaps identification of speaker sentiment, discourse analysis (detecting topic change), lexical analysis (borrowed words), cross-lingual entity recognition, machine translation. Also identification of speech disorders, understanding of cross-cultural sociological factors}

Our contributions are the following:

\begin{itemize}
\item We propose a new task identifying the motivation in Spanish-English code-switching and develop an annotation scheme which identifies 11 different labels, encompassing emotional, pragmatic, and  motivations for code-switching.
\item We create a new dataset applying this annotation scheme to code-switched utterances in the Spanish-English Bangor Miami Corpus ~\cite{deuchar2010bilingbank} and a Hindi-English code-switching dataset consisting of sentences from the GupShup dataset~\cite{mehnaz2021gupshup} and sentences we have written. 
\item We train a baseline Naive Bayes model and fine-tuned the pretrained multilingual models Multilingual BERT~\cite{DBLP:journals/corr/abs-1810-04805} and XLM-RoBERTa~\cite{conneau2019unsupervised} to classify a code-switch with the type(s) of code-switching that it falls into.
\item We conduct a cross-lingual study, transferring the task from Spanish-English code-switching to detection of motivations behind Hindi-English code-switching, thus demonstrating that our annotations can be used cross-lingually. 
\end{itemize}

The outline of the paper is as follows: After reviewing related work (\S\ref{sec:related_work}), we introduce our annotation scheme (\S\ref{sec:annotation}). Next, we conduct experiments with building an automatic system to label code-switched text (\S\ref{sec:automatic}). We then investigate the cross-lingual transfer of our annotations (\S\ref{sec:cross-lingual}), and conclude (\S\ref{sec:conclusion}).

\section{Related Work}
\label{sec:related_work}

Many types of frameworks for code-switching have been created in order to study the motivations behind code-switching~\cite{poplack1980sometimes, gumperz1982discourse, myers1997duelling, zentella1998growing, halim2014functions}, and several studies have annotated code-switched data according to their own frameworks~\cite{lee2015emotion, begum2016functions, hartmann2018integrated, rudra2019identifying}. \newcite{rudra2016understanding} developed classifiers to determine whether Hindi-English code-switching on Twitter was opinionated or not and found that audiences preferred to use Hindi to express a negative sentiment and English to express a positive sentiment. \newcite{lee2015emotion} developed a system to identify the emotions in code-switched Chinese-English posts. Additionally, one corpus of Hindi-English code-switched conversations has broadly grouped the users’ motivations for code-switching in order to study the rules that govern code-switching~\cite{dey2014hindi}. The framework we applied in this paper draws upon elements from \newcite{zentella1998growing}'s framework, and it closely mirrors the approach of \newcite{begum2016functions}. However, while their annotation scheme is based on Tweets, ours is specific to conversational code-switching. 

Previous research has proven the success of fine-tuning the pre-trained models Multilingual BERT and XLM-RoBERTa on the task of identifying offensive language in code-switched texts~\cite{jayanthi2021sj_aj}. However, many factors affect the performance of multilingual language models, which do not always perform higher than monolingual models. \newcite{aguilar2019english} showed that the English model ELMo could be fine-tuned on code-switched data and outperform Multilingual BERT when it came to language identification. \newcite{tang2020fine} found that the BERT-Base Chinese model outperformed the BERT-Base multilingual model for sentiment analysis on Chinese code-switching text because most of their pre-training corpus was in Chinese or English rather than code-switched. Multilingual models do not necessarily have better embeddings than hierarchical meta-embeddings when it comes to code-switching, but XLM-RoBERTa was shown to outperform other models in NER and POS tagging tasks~\cite{winata2021multilingual}. Because of these models’ state-of-the-art performance, we decided to fine-tune Multilingual BERT and XLM-RoBERTa on our tasks.

An emerging field in NLP which has significant relevance for code-switching research is zero-shot cross-lingual transfer learning. Zero-shot cross-lingual transfer learning refers to when a model is trained to perform a task in one language, without any training examples from other languages, then is expected to perform the same task on data from another language which it has never seen before~\cite{srivastava2018zero}. Transformer-based multilingual language models have been shown to have poorer representations of vocabulary in some languages compared to other languages, which reduces their effectivity~\cite{pires2019multilingual}. Training language models to perform well on tasks with code-switched data has been shown to improve their performance on downstream tasks with other languages because it improves models’ generalizability across languages~\cite{qin2020cosda, krishnan2021multilingual}. Still, multilingual language models have been shown to perform poorly in zero-shot cross-lingual transfer learning tasks with low-resource languages~\cite{lauscher2020zero}.

One tool used often in zero-shot cross-lingual transfer learning is cross-lingual word embeddings, which encode words from multiple languages in the same embedding vector~\cite{zhou2020end}. They have also shown promising results with code-switching, outperforming baseline models for voice rendering of code-switched text~\cite{zhou2020end} and semantic parsing~\cite{duong2017multilingual}. While zero-shot cross-lingual transfer learning research with code-switching has largely looked into improvements in performance on monolingual data, to our knowledge, no research has investigated transfering a task in one code-switching language pair to another code-switching language pair.

Machine translation has been applied to code-switching in order to translate a code-switched input to a monolingual language output~\cite{xu2021can}. However, no techniques have been created to translate a code-switch in one language pair to a code-switch in another language pair. We used machine translation through the Google Translate package in order to translate Spanish-English code-switched training data to Hindi-English code-switched data. This method is the Translate-Train strategy in cross-lingual transfer learning, which has the potential to introduce superficial patterns in data and negatively impact model performance~\cite{artetxe2020translation}. 

\section{Annotation}
\label{sec:annotation}

In this section, we describe the data we annotated, discuss the annotation scheme, and give a comparison of our annotation to previous annotation schemes.

\subsection{Data}
We annotated data from the from the Bangor Miami corpus ~\cite{deuchar2010bilingbank}, a publicly available code-switched Spanish-English conversational dataset consisting of audio recordings and transcripts between two or more speakers.
We filtered the data from the transcriptions for sentences with instances of code-switching, included three of the transcript’s preceding and following lines in order to capture sufficient context from the code-switch for later annotation, and preprocessed the data using the Python package re\footnote{https://docs.python.org/3/library/re.html} to remove noise that was annotated. We annotated the first 26 transcripts of the 56 total transcripts. Statistics from our filtered dataset are shown in Table~\ref{table:corpus}. 

\begin{table}[h]
\begin{tabular}{|c|m{2cm}|}
\hline
\bf{Description} & \bf{Count} \\
 \hline 
 Number of utterances & 1,379 \\
 \hline
 Words in Spanish & 15,796 \\
 \hline
 Words in English & 20,357 \\
 \hline
 Ambiguous words & 3,393 \\
 \hline
\end{tabular}
\caption{Statistics of filtered code-switches we used from the Bangor Miami corpus}
\label{table:corpus}
\end{table}

\subsection{Annotation Scheme}
\rmb{ready to edit}
For many decades, code-switching has been studied from the angle of linguistics and sociology. Several researchers, including Ana Celia Zentella, have built frameworks for hybrid languages and categorized types of code-switching. Zentella studied a Spanish-speaking community’s speech over two decades, tracked instances of code-switching, and published a breakdown of the most frequent types of code-switching~\cite{zentella1998growing}.

We drew from~\newcite{zentella1998growing}'s framework on code-switching to identify eleven labels in the annotation scheme as a mix of syntactic and emotional types of code-switching. Like~\newcite{begum2016functions}, we identified that a single code-switch could serve multiple purposes because each code-switch can be seen as a sum of its semantic, structural, and sentiment-related dimensions. Thus, the labels are not mutually exclusive, and one code-switch can have multiple labels.

\textbf{Changing topics} refers to code-switching to introduce another viewpoint, change the tone, or clarify something. Ex: I’m not ready at all, \emph{¿y qué tal tú?} (I'm not ready at all, and what about you?)

\textbf{Borrowing} refers to making a short word or phrase substitution in the other language, then returning to the original language. Ex: \emph{Mi amiga de} high school \emph{va a casarse en dos semanas.} (My friend from high school is going to get married in two weeks.)

\textbf{Making a joke} refers to code-switching for comedic effect or a sarcastic quip. Ex: You’re making such a big deal about it, \emph{como si murieran las personas en la calle.} (You're making such a big deal about it, as if people were dying in the street.) 

\textbf{Quoting} refers to switching languages to be true to how a statement was spoken by someone else. Ex: So my Spanish teacher said, "\emph{Oye, necesitas estudiar más.}" (So my Spanish teacher said, "Hey, you need to study more.")

\textbf{Translating} refers to switching languages to repeat a statement or phrase, perhaps for the sake of emphasis or clarity. Ex: \emph{A veces}, sometimes, I like to be by myself. (Sometimes, sometimes, I like to be myself.)

\textbf{Giving a command} refers to code-switching to make a mandate or imperative intended to get the addressee to do something. Ex: \emph{Él no sabe lo que está diciendo}, just don’t listen to him. (He doesn't know what he's saying, just don't listen to him.)

\textbf{Using a filler} refers to switching languages to use a filler, brief interjection, or short noise intended to communicate meaning from the other language. Ex: \emph{Y yo me callé}, you know, \emph{porque no quería ofender a nadie.} (And I stopped talking, you know, because I didn't want to offend anybody.)

\textbf{Expressing exasperation} refers to code-switching to complain or emphasize anger or frustration. Ex: \emph{Ay, cómo me sigues molestando}, I should just get up and leave! (Oh, how you keep annoying me, I should just get up and leave!)

\textbf{Expressing happiness} refers to code-switching to make a compliment or positive interjection. Ex: I just saw her dress, \emph{¡qué lindo!} (I just saw her dress, how pretty!)

\textbf{Talking about proper nouns} refers to switching languages to talk about people or places whose names are in the other language or pronounced according to the other language. Ex: \emph{Escogimos} United Airlines \emph{porque ellos ofrecen las mejores meriendas.} (We chose United Airlines because they offer the best snacks.)

\textbf{Expressing surprise} refers to code-switching to interject or relay that something was unexpected. Ex: \emph{¿Qué hizo ella?} Oh my god. (What did she do? Oh my god.)

One example of an utterance with more than one type of code-switching is "Once again we're talking about that. Mmhm. And ah this black guy that was doing it was so careless. \textit{Calla} (shut up), don't say that." This counted as both borrowing and giving a command. 61.6\% of the utterances in the dataset contain more than one type of code-switching. It is possible for an utterance to contain code-switching that doesn't fall under our scheme, therefore gets no label, but this does not occur in our dataset. 

\subsection{Comparison to Previous Annotation Schemes}

\rmb{ready to edit}
Our change topic category is closely modeled after \newcite{zentella1998growing}'s designation of Realignment, which includes a topic shift, rhetorical question, break from a narrative, aside comment, and checking with the listener. We share some of the categories that we have annotated with \newcite{begum2016functions}, such as sarcasm (joke), quotations (quote), imperative (command), and translation (translate). Additional categories we have included such as using a filler and expressing happiness, frustration, or surprise are more specific to the type of code-switching that occurs during a conversation in which someone is reacting to the statements made by the other person. Categories that \newcite{begum2016functions} include which we do not are the more fine-grained breakdowns of Narrative-Evaluative, Reinforcement, Cause-Effect, and Reported Speech. \newcite{dey2014hindi} establish a set of motivations for code-switching among the speakers in their Hindi-English code-switching conversation corpus, which consists of Ease of Use, Comment, Referential Function, Topic Shift, Dispreference, Personalisation, Emphasis, No Substitute Word, Name Entity, and Clarification. However, they do not go in depth into their reasoning behind choosing these motivations and offer little elaboration upon what each one entails. \newcite{lee2015emotion}'s annotation scheme for emotions in Chinese-English code-switching includes happiness, sadness, anger, fear, and surprise, three of which we share in our categories of happiness, exasperation, and surprise.

We chose to offer a visual comparison between our annotation scheme, \newcite{zentella1998growing}'s, and \newcite{begum2016functions}'s because of the significant overlap between the three. See Table \ref{annotation comparison} for a full breakdown of the similarities and differences between our annotation systems.\footnote{\newcite{zentella1998growing} also includes the following designations that neither we nor \newcite{begum2016functions} include: future referent check and/or bracket, checking, role shift, rhetorical ask and answer, appositions and/or apposition bracket, accounting for requests, double subject (left dislocation), recycling, triggers, parallelism, and taboos.} 

\begin{table*}[h!]
\begin{center}
\begin{tabular}{ | m{4cm} | l | l | } 
  \hline
  \bf{\newcite{zentella1998growing}} & \bf{\newcite{begum2016functions}} & \bf{Our paper} \\ 
  \hline
  Topic shift, & Narrative-Evaluative, & \multirow{2}{2cm}{Change topic} \\ 
  Declarative/question shift & Cause-Effect &  \\
  \hline
  Narrative frame break 
  & Sarcasm & Joke \\
  \hline
  Direct Quotations & Quotations & \multirow{2}{2cm}{Quote} \\
  \cline{1-2}
  Indirect Quotations & Reported Speech & \\
  \hline
  Aggravating requests & \multirow{2}{3cm}{Imperative} & \multirow{2}{2cm}{Command} \\
  \cline{1-1}
  Mitigating requests & & \\
  \cline{1-1}
  Attention attraction & & \\
  \hline
  Translations & Translation & Translate \\
  \hline
  - & Reinforcement & - \\
  \hline
  Crutching & - & Borrowing \\
  \hline
  Filling in & - & Filler \\
  \hline
  - & - & Proper noun \\
  \hline
  -  & - & Happiness \\
  \hline
  - & Abuse/Neg. Sentiment & Exasperation \\
  \hline
  - & - & Surprise \\
  \hline
\end{tabular}
\caption{Comparison of our annotation scheme with other frameworks for code-switching}
\label{annotation comparison}
\end{center}
\end{table*}


\subsection{Inter-Annotator Agreement}

A subset of 100 code-switched utterances were labeled by another annotator and the inter-annotator reliability score was calculated using accuracy against the principal annotator. The agreement scores are shown in Table \ref{table:accuracy agreement} for each category.

\subsection{Statistics and Observations}

In this subsection, we discuss some statistics from the dataset. In the annotated data, the frequency of some types of code-switching over others validates theories about code-switching. For example, code-switching to change topics is regarded as the most frequent type of code-switching \cite{zentella1998growing}, a trend which is present in Table \ref{table:label_distribution}. There were three entries which had been filtered that contained markers that a code-switch was near, but were all spoken in one language, so they received no label.

\begin{table}[h]
\begin{tabular}{|c|c|m{2.7cm}|}
 \hline 
 \bf{Label} & \bf{Frequency} & \bf Inter-annotator Accuracy \\
 \hline
 Change topic & 65.0\% & 82\% \\ 
 \hline
 Borrowing & 26.0\% & 78\% \\
 \hline
 Joke & 3.7\% & 91\% \\
 \hline
 Quote & 6.5\% & 95\% \\
 \hline
 Translate & 5.9\% & 92\%\\
 \hline
 Command & 8.3\% & 9.0\% \\
 \hline
 Filler & 31.0 \% & 77\% \\
 \hline
 Exasperation & 7.7\% & 90\% \\
 \hline
 Happiness & 4.1\% & 94\% \\
 \hline
 Proper noun & 25.9\% & 88\% \\
 \hline
 Surprise & 11.6\% & 80\% \\
 \hline
\end{tabular}
\caption{Distribution of labels in the dataset (Frequency) and accuracy of the agreement between annotators (Inter-annotator Accuracy).}
\label{table:label_distribution}
\label{table:accuracy agreement}
\end{table}

\section{Automatic Detection of the Type of Code-Switching in Conversations}
\label{sec:automatic}
\rmb{ready to edit, except where noted}

We trained classifiers on our annotated corpus to predict labels for the code-switching text automatically. Results show the most effective approach is by building unique classifiers for each label and the best performance is typically achieved by the XLM-RoBERTa model with an adapter layer.

In addition to a baseline Naive Bayes classifier, we also fine-tuned bert-base-multilingual-cased (mBERT) and xlm-roberta-base (XLM-RoBERTa) classifiers, using the Huggingface library\footnote{https://huggingface.co/models. mBERT base has 110M parameters, and XLM-RoBERTa base has 125M. We used the Google Colab Pro+ Tesla V100-SXM2-16GB GPU to train the models, and each model trained in less than 15 minutes.}. We chose these multilingual models as they have been trained on multilingual corpora and thus, are better equipped to classify our multilingual data. Because of the relatively small training set, to combat overfitting, we experimented with adapter layers to the two Transformers. Adapter layers decrease the number of trained parameters by adding a small number to tunable parameters and keeping the rest of the model fixed.

\subsection{Experimental Setup}

Four conversations (16\% of the annotated data, 220 code-switches) were randomly set aside as test data, and the rest of the data was organized into a 75/25 train/dev split. For the mBERT and XLM-RoBERTa models as well as their respective adapter models, our hyperparameters were 20 epochs, a weight decay of 0.01, and we tuned the batch size from the set ${4, 16}$ and the learning rate from the set ${2e-05, 0.0001}$ with grid search. In order to account for the variance between different initial seeds, we first found the best performing hyperparameter combination for each model on each task with the default seed of 42, then we ran the model four additional times with the same hyperparameters but with a different seed, from 30 to 20 to 10 to 5.

\begin{table*}[h!]
\begin{center}
\begin{tabular}{ | m{2.5cm} | m{1cm}| m{2cm} |  m{2cm} | m{2cm} | m{2cm} | } 
  \hline
  \bf{Label} & \bf{Naive Bayes} & \bf{mBERT} & \bf{mBERT with adapter} & \bf{XLM-R} & \bf{XLM-R with adapter}\\ 
  \hline
  Change topic & 63.2 & \bf{86.3} $\pm 1$ & 85.7 $\pm 1.7$ & \bf{86.3} $\pm 0.9$ & \bf{86.3 $\pm 0.4$} \\ 
    \hline
    Borrowing&57.3 & \bf{78.5} $\pm 6.7$& 77.4 $\pm 3.1$& 75 $\pm 2.3$& 70.9 $\pm 2.1$ \\ 
    \hline
  Joke & 59.6 & \bf{79.8 $\pm 13.6$} & 37.0 $\pm 28.0$ & 68.5 $\pm 15.6$& 68.7 $\pm 9.8$  \\ 
  \hline
  Quote & 40.9 & \bf{75.6 $\pm 2.4$} & 74.3 $\pm 5.2$ & 69.3 $\pm 4.9$ &
70.3 $\pm 4.6$ \\
  \hline
  Translate&
46.4 & 72.2 $\pm 10.7$&\ 73.9 $\pm 9.6$ & \bf{74.6} $\pm 17.6$& 74 $\pm 10.5$ \\ 
    \hline
  Command&70.5 & 59.6 $\pm 31$ & \bf{74.3} $\pm 8.2$ & 66.4 $\pm 20.6$ & 66.2 $\pm 7.1$\\ 
    \hline
    Filler & 57.8 & 70.5 $\pm 3.2$ & 72.2 $\pm 5.3$ & 73.4 $\pm 2.5$ & \bf{74.4} $\pm 2.5$ \\
    \hline
  Exasperation&62.3 & 53.2 $\pm 16.8$ & 51.4 $\pm 14.2$ & 70.5 $\pm 14.4$& \bf{77.1} $\pm 8.7$ \\
      \hline
  Happiness&64.1 &\bf{83.6} $\pm 6.1$ & 80.2 $\pm 8.7$ & 78.4 $\pm 4.3$ &70.5 $\pm 6.3$\\ 
      \hline
  Proper noun&61.0 & 84.5 $\pm 3.3$ & 85.4 $\pm 1.6$ & \bf{85.5} $\pm 1.9$&83.6 $\pm 1.9$ \\ 
  \hline
  Surprise&68.2 & 75.0 $\pm 4.9$ & 66.4 $\pm 3.9$ & \bf{79.4} $\pm 3.6$& 73.3 $\pm 7.4$ \\ 
  \hline
  \bf{Average}& 59.2 & 74.4 $\pm 2.8$ & 70.7 $\pm 5.2$& \bf{75.4} $\pm 3.6$ & 74.1 $\pm 3.1$ \\ 
  \hline
\end{tabular}
\caption{Accuracy (in \%) of label detection in code-switching dialogue. We report the standard deviation from training with 5 different random seeds.}
\label{accuracy spanglish}
\end{center}
\end{table*}

\subsection{Results}
The accuracy for each label with each model is shown in Table \ref{accuracy spanglish}.
We compute the mean accuracy of each model on each task and reported the standard deviation across training runs in order to understand the statistical significance in the difference of the models' performance. We find that there is no one best model, but that different deep learning models perform better on some labels.  Overall, the best model is XLM-RoBERTa with an adapter layer.

\subsection{Qualitative Analysis of Results}
\rmb{ready to edit}

We looked at examples of the models' predictions and have included some in Table \ref{qualitative spanglish}. In the first example, the model correctly predicted borrowing because of the English word "trainers" surrounded by Spanish. This was a difficult example because of the potential noise introduced by the earlier code-switch for changing topics. We believe that the borrowed word was surrounded by enough Spanish tokens to be a noticeable code-switch. In the second example, the model missed the code-switch for using a filler, which was "\emph{bueno}." We believe that when there are multiple code-switching points, as there were in this example, it is difficult for the model to identify one word that indicates a filler. In the third example, the model correctly predicted translations, which came from the phrase "my dad was (just) the same" being said once in English, then later in Spanish. This was also a difficult example because the translation code-switch was not said immediately, but rather a few statements afterwards, but the model was still successful in identifying the translation. In the fourth example, the model mistakenedly predicted giving a command. We believe that it picked up on the last word, \emph{escúpame} (spit me), as a command, but the model missed that this word was being used in a quoting capacity, not in the literal sense. 

\begin{table*}[h!]
\begin{center}
\begin{tabular}{ | m{11cm} | l | m{1.8cm} | } 
  \hline
  \bf{Transcript} & \bf{Gold} & \bf{System} \\ 
  \hline
  \textbf{Original:} MAR: That my children were being welcomed into the — \emph{olvídate si tiene como tres} trainers! \emph{Tiene un cocinero!} & Borrowing & Borrowing \\
  \cline{1-1}
  \textbf{Translation:} MAR: That my children were being welcomed into the — forget it if he has like three trainers! He has a chef! & & \\
  \hline
  \textbf{Original:} JES: \emph{Invita a a alguna de las celebraciones.} NIC: I don't know. I have JES: \emph{Tú sabes se caen bien.} NIC: Yeah I'll tell her. \emph{Bueno} not her I gotta tell sister. & Filler & No Filler \\
  \cline{1-1}
  \textbf{Translation:} JES: Invite [her] to one of the celebrations. NIC: I don't know. I have JES: You know they like each other. NIC: Yeah I'll tell her. Well not her I gotta tell sister. & & \\
  \hline
  \textbf{Original:} IRI: \emph{Ajá}. JAM: If if I happens to see like running blood or something like that I feel disgust and I feel weak and I IRI: My dad was just the same. \emph{Sí, sí, sí, sí}, kryptonite. JAM: Kryptonite, yeah. IRI: \emph{No mi pa mi papá era igual.} & Translate & Translate \\
  \cline{1-1}
  \textbf{Translation:} IRI: Uh huh. JAM: If if I happens to see like running blood or something like that I feel disgust and I feel weak and I IRI: My dad was just the same. Yes, yes, yes, yes, kryptonite. JAM: Kryptonite, yeah. IRI: No my da- my dad was the same. & & \\
  \hline
  \textbf{Original:} PAI: \emph{En qué lo puedo ayudar?} SAR: He's going to the airport. PAI: What up. SAR: \emph{Discúlpame}. PAI: It sounds like you're saying \emph{escúpame}. & Command & No \break Command \\
  \cline{1-1}
  \textbf{Translation:} PAI: How can I help you? SAR: He's going to the airport. PAI: What up. SAR: Excuse me. PAI: It sounds like you're saying spit me. & & \\
  \hline
\end{tabular}
\caption{Sample system outputs on Spanish-English code-switched data with speaker IDS\jmf{add speaker id}}. We show gold and system outputs for only one label type.  However, these examples may have additional other labels. \rmb{resolved - is it clarified that we don't have speaker ids fed into the model?}
\label{qualitative spanglish}
\end{center}
\end{table*}

\begin{table*}[h!]
\begin{center}
\begin{tabular}{ | m{11cm} | l | m{1.8cm} | } 
  \hline
  \bf{Transcript} & \bf{Gold} & \bf{System} \\ 
  \hline
  \textbf{Original:} ASH: At seven into work JAC: You have to go with me, at seven? ASH: At seven. JAC: \emph{Aah.} ASH: \emph{Main bahut utsaahit tha main khush, khush tha} & Happiness & Happiness \\
  \cline{1-1}
  \textbf{Translation:} ASH: At seven into work JAC: You have to go with me, at seven? ASH: At seven. JAC: Ah. ASH: I was so excited I was happy, happy. & &\\
  \hline
  \textbf{Original:} MAR: \emph{Yah havaee adde ke paas hai. Ok, unhen kaha jaata hai a} brokers. \emph{Mainne use isalie nahin bulaaya tha kyonki} Pedro \emph{kaha tha.} & No Quote & Quote \\
  \cline{1-1}
  \textbf{Translation:} MAR: It's near the airport. Ok, they're called brokers. I didn't call him because Pedro said. & & \\
  \hline
\end{tabular}
\caption{Sample system outputs on Hindi-English code-switched data with speaker IDs. We show gold and system outputs for only one label type.  However, these examples may have additional other labels.}
\label{qualitative hinglish}
\end{center}
\end{table*}

\section{Cross-Lingual Transfer}
\label{sec:cross-lingual}

\rmb{ready to edit, except for results and table}
In this section, we describe the second phase of our paper, which was the cross-lingual transfer of the task of identifying the motivations behind code-switching from Spanish-English code-switching to Hindi-English code-switching.

\subsection{Data}

In searching for the appropriate Hindi-English code-switched dataset for this task, we considered the need for a conversational dataset in order to be consistent with the system we had created. We found that the GupShup dataset~\cite{mehnaz2021gupshup}, which contains over 6,800 Hindi-English conversations, was in the similar domain as the Bangor Miami corpus. One downside was that dataset was synthetic: the sentences had been translated from the SAMSum corpus (written in monolingual English) to a Hindi-English code-switched format~\cite{mehnaz2021gupshup}. However, we decided to use the dataset on the basis of its close replication of natural Hindi-English code-switching. The original dataset had been written by linguists to mimic natural conversations, and the GupShup translation was done by fluent speakers of Hindi and English who used code-switching in their day-to-day lives, so we decided to use the dataset~\cite{mehnaz2021gupshup}.\footnote{The GupShup dataset is not publicly available, and in order to use it, we abided by the terms of not sharing it with others and only using it for our research purposes. We believe our use is consistent with the intended use of the dataset.} In addition to using code-switching examples from the GupShup sentences, we also added our own Hindi-English code-switched sentences in order to get a wider range of communicative phenomenon, with 150 total Hindi-English code-switched sentences in the test data.

\subsection{Methods}

We translated our training and validation data from Spanish-English code-switching into Hindi-English code-switching. First, we used the Facebook AI library fasttext\footnote{https://github.com/facebookresearch/fastText} to detect the language of each word in the annotated Bangor Miami training and validation data for each task. Once each word had been labeled as either Spanish or English, we applied the Google Translate Python library\footnote{https://pypi.org/project/googletrans/} to translate the Spanish segments of the Spanish-English code-switching into transliterated Hindi. 

Out of the best hyperparameter training combination for each model on the Spanish-English code-switching data, we picked the model which performed the best on each respective task and trained it on the translated Hindi-English code-switching data. Then, for each task, we tested the model using the test Hindi-English code-switched data.

\subsection{Results}
\jmf{todo for jeff}
\rmb{Say what the table says, with some reasons, we think these are lower because xyz }
The accuracy of each task on the Hindi-English code-switched data is shown in the table below, as well as the overall accuracy of all tasks. We find that the cross-lingual applicability of our annotation scheme is quite high, achieving 66\% accuracy overall.

\rmb{Add table once all code-switching examples have been created and models have been run}

\begin{table}[h!]
\begin{center}
\begin{tabular}{ | m{2.5cm} | m{2.3cm} | } 
  \hline
  \bf Label & \bf Accuracy (\%) \\ 
  \hline
  Change topic & 75.9 $\pm 2.9$ \\
  \hline
  Borrowing & 76.3 $\pm 2.4$ \\
  \hline
  Joke & 78.1 $\pm 12.9$ \\
  \hline
  Quote & 72.6 $\pm 11.8$ \\
  \hline
  Translate & 37.4 $\pm 6.3$ \\
  \hline
  Command & 46.7 $\pm 12.9$ \\
  \hline
  Filler & 81.5 $\pm 2.3$ \\
  \hline
  Exasperation & 66.3 $\pm 10.8$ \\
  \hline
  Happiness & 50.7 $\pm 8.4$ \\
  \hline
  Proper noun & 84.8 $\pm 5.8$ \\
  \hline
  Surprise & 57.8 $\pm 13.5$\\
  \hline
  Average & 66.2 $\pm 2.1$ \\
  \hline
\end{tabular}
\caption{Accuracy (in \%) of label detection in Hindi-English code-switching dialogue}
\label{accuracy hinglish}
\end{center}
\end{table}

\subsection{Qualitative Analysis of Results}

We include some examples of the models' predictions in Table \ref{qualitative hinglish}. In the first example,the model correctly predicts the code-switch as an example of happiness, as the speaker transitions to make a happy exclamation. In the second example, the model mistakenly predicts quoting. We believe that the model was triggered by the Hindi word translating to "said," which commonly appears in examples with quoting.

\section{Conclusion}
\label{sec:conclusion}

\rmb{ready to edit}
This paper presents a corpus of Spanish and English code-switching with labels for the different motivations for code-switching. We collected the data from the Bangor Miami corpus, created an annotation scheme for types of code-switching, and annotate the data. We proposed a classifier-based approach to detect the types of code-switching in the annotated code-switching corpus. Results show that the XLM-RoBERTa model is the most effective at predicting types of code-switching, and that our annotations are applicable to new language pairs. We believe that analysis of types of code-switching is an innovative approach towards bilingual speech diagnosis as well as contributing to a linguistic model of code-switching.

\textbf{Limitations}: Our system has been trained on everyday conversations from Spanish-English bilinguals, and may not be applicable to other languages or domains. There is a risk that incorrect conclusions can be drawn if the system does not meet the performance requirements.


\bibliography{anthology,custom}
\bibliographystyle{acl_natbib}

\end{document}